\documentclass{article}

\usepackage{PRIMEarxiv}

\usepackage[utf8]{inputenc} 
\usepackage[T1]{fontenc}    
\usepackage{hyperref}       
\usepackage{url}            
\usepackage{booktabs}       
\usepackage{amsfonts}       
\usepackage{nicefrac}       
\usepackage{microtype}      
\usepackage{lipsum}
\usepackage{fancyhdr}       
\usepackage{graphicx}       
\graphicspath{{media/}}     

\title{InceptionTime vs. Wavelet – a comparison for time series classification
}

\author{
  Daniel Klenkert, Daniel Schaeffer, Julian Stauch \\
  Technology Campus Teisnach Sensor Technology \\
  Deggendorf Institute of Technology \\
  Teisnach, 94244 Germany\\
  \texttt{daniel.klenkert@th-deg.de} \\
}

\begin{document}
\maketitle

\begin{abstract}
Neural networks were used to classify infrasound data. Two different approaches were compared. One based on the direct classification of time series data, using a custom implementation of the InceptionTime network. For the other approach, we generated 2D images of the wavelet transformation of the signals, which were subsequently classified using a ResNet implementation.
Choosing appropriate hyperparameter settings, both achieve a classification accuracy of above 90 \%, with the direct approach reaching 95.2 \%.
\end{abstract}

\keywords{classification \and transfer learning \and time series data \and wavelet}

\section{Introduction}
Infrasound carries a lot of information about natural, as well as man-made events. Since infrasound waves can travel for long distances in earth’s atmosphere, they are ideally suited to detect strong atmospheric events with a sparse global network of measuring stations. Such a network was realized as part of the international monitoring system (IMS), with the primary goal of detecting nuclear explosions \cite{1}. However other infrasound sources, for example, earthquakes, meteorites and ocean waves are also detected with this system.

Therefore, it is necessary to distinguish between different sources of infrasound. Artificial intelligence methods, especially neural networks, offer a suitable approach for this classification of infrasound data. Prior to using any neural network, it is necessary to train it. For this reason, labeled training data was used, which was generated by numerical simulation of eight different classes of infrasound-producing events.

\section{General Approach}
\label{sec:Approach}

For the data preprocessing, model training and model evaluation we took advantage of existing high-level API libraries available in Python. In particular, we used the fastai \cite{2} and tsai libraries \cite{3} extensively. Both libraries support GPU utilization significantly speeding up the training process compared to pure CPU calculations. Additionally, they also provide pre-trained neural networks, which we used for transfer learning to reduce the number of calculations necessary to train the network.

Two fundamentally different classification approaches were explored:

\textbf{Direct approach}: 

The simulated signals are directly fed into an appropriate neural network as time series data.

\textbf{Wavelet approach}: 

The signals are transformed into 2D spectral data using a wavelet transformation \cite{4}. This spectral data can be plotted as a heatmap image, which is subsequently classified using standard image classification techniques. Figure 1 shows two examples of wavelet-transformed time series out of the training set. The images are then classified using a neural network optimized for image recognition.

\begin{figure}
  \centering
  \includegraphics[width=0.8\textwidth]{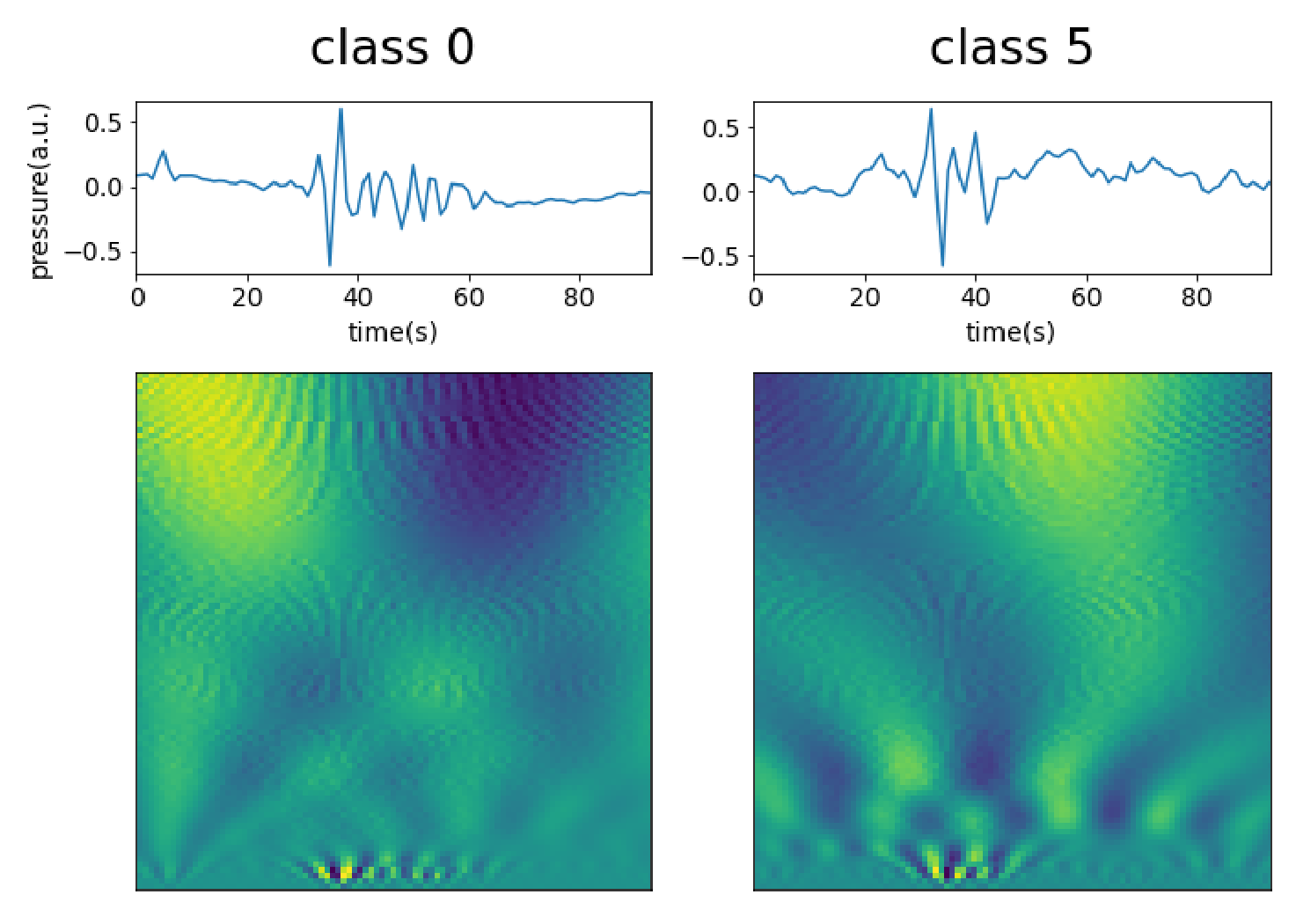}
  \caption{Raw signals and corresponding wavelet transformation of two different time series signals out of category 0 and 5.}
  \label{fig:fig1}
\end{figure}

\section{Data Preprocessing}
\label{sec:Preprocessing}
The training data consisted of 2400 different infrasound signals, containing 94 data points each. To gauge the accuracy of the model and detect possible overfitting in the training process, a validation set was randomly chosen from the 2400 signals. To ensure comparability between different training runs, the random splitting was always initialized with the same seed. This set includes 480 (i.e. 20 \%) of the total signals available. The training of the neural network only takes place on the remaining 1920 signals (the training set), so the validation set can be used to independently judge the accuracy of the trained model.

For the direct approach, both training and validation set are directly used to construct a TSDataLoaders object, which is a class provided by tsai. This object is used to feed the training data into the neural network for training. TSDataLoaders groups both data sets into batches of 64 signals and standardizes each batch according to the mean and standard deviation of the data points it contains. 

For the wavelet approach, a wavelet transformation is applied to each signal, creating a 2D data object from the 1D signal. This 2D data object is visualized as a heatmap image. The images created from all the signals are then used to create an ImageDataLoaders object, which is the TSDataLoaders equivalent for image data in fastai.

\section{Model Development}
\label{sec:Model}
\textbf{Direct approach:}

We tested different models, like MiniRocket \cite{5} for our neural network. In the end, we chose InceptionTime \cite{6} because of its good performance. This network was specially developed for tasks focused on time series classification and it is natively supported by tsai. To this neural network a custom output layer was added, to reduce the number of output neurons from 128 in the original model to eight, which corresponds to the number of classes of events, which are to be identified.

\textbf{Wavelet approach:}

In the wavelet approach, the problem of classifying a time series signal is transformed into classifying an image. For this task, different versions of the ResNet neural network are known to produce excellent results. As they are natively available in fastai, ResNet models of different sizes were used to classify the spectral images.

The training data is fed into the neural network using TS/ImageDataLoaders, as explained in the previous chapter. Regardless of the specific approach, the model is trained for a certain number of epochs with the training set. After each epoch, the model is used to predict the classes of the signals in the validation set to gauge its accuracy. However, no adjustment of the weights and biases are conducted with the results from evaluating the validation set. 

\section{Results}
\label{sec:Results}

\textbf{Direct approach:}

We focused on finetuning the InceptionTime model using the fastai library. Starting with a learning rate of $1e-3$ and 20 epochs, we could already achieve up to 93 \% accuracy. 
Raising the number of epochs to 50 showed no further improvement. So, the next step was improving the learning rate. This was done by using $lr_{find}$ and some manual testing. Repeating the training with this new learning rate $lr_{max}=(slice(1e-6, 1e-2))$, 50 epochs and monitoring the accuracy, showed 33 epochs works best. Thus, a final classification accuracy of 95.2 \% was scored on the validation data set.

\textbf{Wavelet approach:}

In this approach, different parameters can be tuned. First, there is a set of parameters for the wavelet transformation, including the applied wavelet, the frequency band of the transformation and the colormap for the image generation. It turned out, that using the full available frequency band (i.e.  the number of frequency steps equals the number of time steps in the original signal) works best. With the standard viridis colormap and the morlet wavelet the best results could be achieved.
For the training of the ResNet50 for the image classification a learning rate of $lr_{max}=2e-2$ has shown the best performance with a final accuracy of 90.2 \%.

\section{Conclusion and future work}
In conclusion the direct approach performed slightly better in terms of accuracy. It also significantly outperformed the wavelet approach on training speed, since in the direct approach only 94 datapoints had to be processed for each signal, compared to $94^2=8836$ datapoints in the wavelet approach. Due to these advantages, the direct approach with the InceptionTime network shows a better performance.
Concerning the splits between training and validation sets, only random splits were used to ensure no underlying systematic trend in the data is exploited. Since training a neural network has some randomness to it, the measured accuracy can fluctuate within a range of a few percentage points, independent of the chosen approach. To reduce these fluctuations, a random seed was set for the splitting between the training and validation sets.

The described approach to classifying infrasound signals can easily be generalized to the classification of arbitrary time-dependent signals. Such signals occur in a variety of different scientific and engineering fields. For example, electric signals measured with an oscilloscope or the vibration of machines and buildings.
AI methods, like time-series classification could be used for predictive maintenance and a whole range of other applications. 
To ensure a faster proliferation of AI methods to different application domains it is critically important to make AI more accessible to people, who are not experts in informatics. High-level libraries like the ones we used in this project show the way in this direction.

\bibliographystyle{unsrt}  
\bibliography{references}

\end{document}